\documentclass[10pt,twocolumn,letterpaper]{article}

\usepackage{cvpr}              

\usepackage{graphicx}
\usepackage{amsmath}
\usepackage{amssymb}
\usepackage{booktabs}
\usepackage{bbold}

\usepackage[pagebackref,breaklinks,colorlinks]{hyperref}

\usepackage[capitalize]{cleveref}
\crefname{section}{Sec.}{Secs.}
\Crefname{section}{Section}{Sections}
\Crefname{table}{Table}{Tables}
\crefname{table}{Tab.}{Tabs.}

\def\cvprPaperID{9205} 
\def\confName{CVPR}
\def\confYear{2022}

\usepackage{overpic}
\usepackage{enumitem} 
\usepackage{overpic} 
\usepackage{color}

\definecolor{turquoise}{cmyk}{0.65,0,0.1,0.3}
\definecolor{purple}{rgb}{0.65,0,0.65}
\definecolor{dark_green}{rgb}{0, 0.5, 0}
\definecolor{orange}{rgb}{0.8, 0.6, 0.2}
\definecolor{red}{rgb}{0.8, 0.2, 0.2}
\definecolor{darkred}{rgb}{0.6, 0.1, 0.05}
\definecolor{blueish}{rgb}{0.0, 0.3, .6}
\definecolor{light_gray}{rgb}{0.7, 0.7, .7}
\definecolor{pink}{rgb}{1, 0, 1}
\definecolor{greyblue}{rgb}{0.25, 0.25, 1}

\usepackage{blindtext}

\renewcommand{\paragraph}[1]{\vspace{1em}\noindent\textbf{#1}.}
\begin{document}
\title{Pair DETR: Contrastive Learning Speeds Up DETR Training}

\author{Seyed Mehdi Iranmanesh\\
Amazon \\
{\tt\small mehdiir@amazon.com}
 \and
 Xiaotong Chen\\
 University of California, Santa Barbara\\
 {\tt\small xchen774@umail.ucsb.edu}
  \and
 Kuo-Chin Lien\\
 Appen\\
 {\tt\small klien@appen.com}
}
\maketitle
\begin{abstract}
The DETR object detection approach applies the transformer encoder and decoder architecture to detect objects and achieves promising performance. In this paper, we present a simple approach to address the main problem of DETR, the slow convergence, by using representation learning technique. In this approach, we detect an object bounding box as a pair of keypoints, the top-left corner and the center, using two decoders. By detecting objects as paired keypoints, the model builds up a joint classification and pair association on the output queries from two decoders. For the pair association we propose utilizing contrastive self-supervised learning algorithm without requiring specialized architecture. Experimental results on MS COCO dataset show that Pair DETR can converge at least 10x faster than original DETR and 1.5x faster than Conditional DETR during training, while having consistently higher Average Precision scores.
\end{abstract}
\section{Introduction}
 Object detection utilizing transformers (DETR)~\cite{carion2020end} is a recent approach that represents object detection as a direct prediction problem via a transformer encoder-decoder~\cite{vaswani2017attention}. DETR reaches a competitive performance with Faster R-CNN~\cite{ren2015faster} without using hand-designed sample selection and non-maximum suppression (NMS). However, DETR approach suffers from slow convergence on training, which needs large-scale training data and long training epochs to get good performance.
 
\begin{figure}
\centering
\begin{tabular}{cc} 
\includegraphics[height=0.7\columnwidth,keepaspectratio]{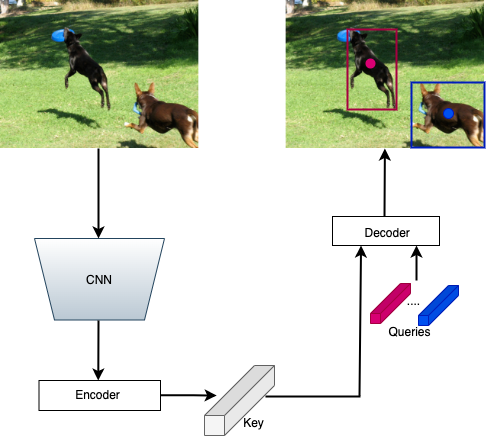} \\ (a) \\ \includegraphics[height=0.7\columnwidth,keepaspectratio]{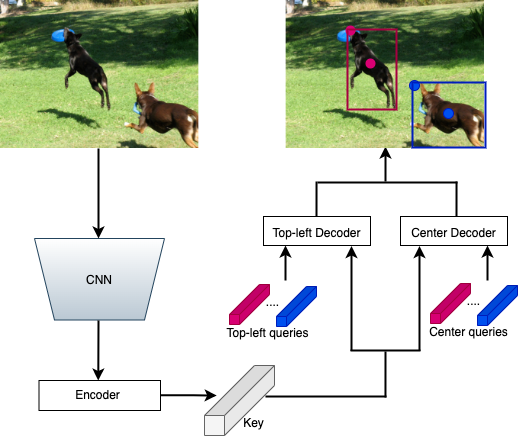} \\
(b)
\end{tabular}
\caption{Transformer based detection models such as DETR~\cite{carion2020end} and Deformable DETR~\cite{zhu2020deformable} rely only on detecting center points (\ref{fig:introduction}a). Pair DETR proposes to detect additional point namely Top-left to localize objects more effectively (\ref{fig:introduction}b).  }
\label{fig:introduction}
\end{figure}

DETR~\cite{carion2020end} uses a set of learned object queries to reason about the associations of the objects and the global image context to output the final predictions set. However, the learned object query is very hard to explain. It does not have a distinct meaning and the corresponding prediction slots of each object query do not have a unique mode.

Deformable DETR~\cite{zhu2020deformable}, handles the issues of DETR by replacing the global dense attention (self-attention and cross-attention) with deformable attention that attends to a small set of key sampling points and using the high-resolution and multi-scale encoder. This however reduced their AP on the bigger objects which needs global dense attention.
 
Conditional DETR~\cite{meng2021conditional} approach is motivated by high dependence on content embeddings and minor contributions made by the spatial embeddings in cross-attention mechanism in DETR. The experimental results in DETR ~\cite{carion2020end} illustrate that if removing the positional embeddings in keys and the object queries from the decoder layer and only using the content embeddings in keys and queries, the detection AP drops. The AP reduction is more observable when the positional embeddings are dropped from the decoder compared the the case when they are dropped from the encoder. 

Conditional DETR approach, learns a conditional spatial embedding for each query from the corresponding previous decoder output embedding. It shrinks the spatial range for the content queries to localize the effective regions for class and box prediction. As a result, the dependence on the content embeddings is less and the training is easier.

Inspired by the great success of contrastive learning and self-supervised training in different applications such as CLIP~\cite{radford2021learning} and SimCLR~\cite{chen2020simple}, we aim to utilize it in transformer training of DETR.
 
Two corner points, e.g. the top-left and bottom-right corners, can define the spatial extent of a bounding box, providing an alternative to the typical 4-d descriptor consisting of the box’s center point and size which is also used in DETR. This has been used in several bottom-up object detection methods such as~\cite{law2018cornernet,zhou2019bottom,tychsen2017denet} and has shown significantly better performance in object localization but not object classification~\cite{chen2020reppoints}.

In order to take advantage of a better localization we propose Pair DETR which consists of a pair of decoders, namely Center decoder and Top-left decoder. Center decoder is responsible to detect the center of bounding boxes while top-left decoder task is to predict top-left of the corresponding bounding boxes (see Fig.\ref{fig:introduction}). In addition to the center prediction, the center decoder is responsible for class prediction and the top-left decoder obeys the classification decision. The learned object queries are shared between two decoders, and as a result the order of center and top-left outputs are the same. To have a more discriminative embedding, contrastive learning approach is utilize between these two sets of output. The contributions of the paper are summarized as follows:

\begin{itemize}
\item We propose to use contrastive learning on features of two decoders to improve the embedding subspace of decoder and improve the classification performance. 
\item We incorporate an extra keypoint \emph{Top-left} and associate it with the the standard keypoint -- center of a bounding box. This novel two-point Transformer architecture improves the localization performance. 
\item  Extensive experiments are conducted to prove the effectiveness of the proposed model comparing to state-of-the-art on COCO dataset. 
\end{itemize}

%


\begin{figure*}
\centering
\includegraphics[width=0.6\textwidth]{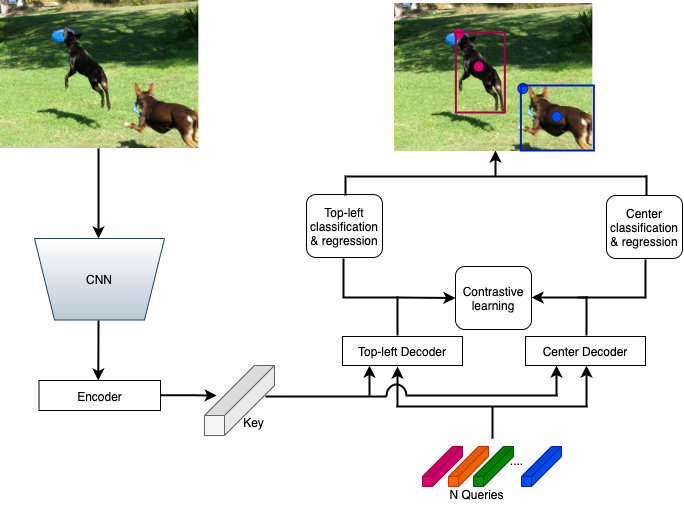}
\caption{\textbf{Pair DETR} contains additional decoder comparing to the other DETR based detection models. One decoder is dedicated to detection of center points (Center decoder) and the other is dedicated to locating top-left points (Top-left decoder). Pair DETR is optimized jointly using classification and regression losses for each decoder and contrastive loss between the two mentioned decoders.  }
\label{fig:method_overall}
\end{figure*} 

\section{Related works}
\subsection{Convolution based object detection}

Most object detection models make predictions from carefully designed initial guesses. Anchor boxes or object centers are two main approaches for making the initial guesses. Anchor based solutions which borrow the idea from the proposal based models, can be categorized into two-stage models such as Faster RCNN~\cite{ren2015faster} variants, and one-stage approaches such as SSD~\cite{liu2016ssd}, and YOLO~\cite{redmon2016you}. 

Two-stage detectors generate a sparse set of regions of interest (RoIs) and classify each of them by a network. Faster RCNN~\cite{ren2015faster} introduced a region proposal network (RPN), which generates proposals from a set of pre-determined candidate boxes, namely anchor boxes. R-FCN~\cite{dai2016r} further improves the efficiency of Faster-RCNN by replacing the fully connected sub-detection network with a fully convolutional sub-detection network. DeNet~\cite{tychsen2017denet} is a two-stage detector which generates RoIs without using anchor boxes. It determines how likely each location belongs to either the  top-right, top-left, bottom-right, or bottom-left corner of a bounding box. It then creates RoIs by enumerating all possible corner combinations, and performs the standard two-stage approach to classify each RoI.

One-stage detectors such as SSD and YOLO, remove the RoI pooling step and detects objects in a single network. One-stage detectors are usually more computationally efficient than two-stage detectors while maintaining comparable performance on different challenging benchmarks. SSD  directly classifies and refines anchor boxes that are densely defined over feature maps at different scales. YOLOv1~\cite{redmon2016you} predicts bounding boxes directly from an image, and later is improved in YOLOv2~\cite{redmon2017yolo9000} by utilizing anchors.

Anchor-free models such as YOLOv1, instead of using anchor boxes, predict bounding boxes at points near the center of objects. Only the points near the center are used since they are considered to be able to produce higher quality detection. YOLOv1 struggles with low recall, and thus YOLOv2 chooses to go back employing anchor boxes. Compared to YOLOv1, FCOS~\cite{tian2019fcos} can utilize all points in a ground truth bounding box to predict the bounding boxes and make comparable results to anchor based solutions. CenterNet~\cite{zhou2019objects} predicts the center, width, and height of objects with hourglass networks, demonstrating promising performance. CornerNet~\cite{law2018cornernet} is another keypoint-based approach, which directly detects an object using a pair of corners and groups them to form the final bounding box. 

\subsection{Transformer based object detection}

DETR successfully applies Transformers to object detection, effectively removing the need for many hand-designed components like non-maximum suppression or anchor generation. The high computation complexity issue, caused by the global encoder self attention, is handled by sparse attentions in Deformable DETR~\cite{zhu2020deformable} or by utilizing adaptive clustering transformer~\cite{zheng2020end}. Deformable DETR~\cite{zhu2020deformable} is introduced to restrict each object query to attend to a small set of key sampling points around the reference points, instead of all points in feature map. Using multi-scaling and utilizing high resolution feature maps, it can mitigate potential losses in the detection of small objects.

The other critical issue, slow training convergence in DETR~\cite{carion2020end}, has been attracting a lot of recent research attention. Deformable DETR~\cite{zhu2020deformable} adopts deformable attention to replace decoder cross-attention. TSP approach~\cite{sun2021rethinking} combines the FCOS and R-CNN-like detection heads and removes the cross-attention modules. With unsupervised pre-training of the transformers, UP-DETR~\cite{dai2021up} enhances DETR performance on object detection precision and also speed of model convergence. Conditional DETR~\cite{meng2021conditional} encodes the reference point as the query position embedding. It utilizes the reference point to generate position embedding as the conditional spatial embedding in the cross-attention layers.  Using spatial queries shrinks the spatial range for the content queries to localize the regions for box and class prediction. Therefore, the dependence on the content embeddings is relaxed and the training convergence is faster. Our proposed approach is based on conditional DETR to improve the performance of DETR object detection while taking advantage of conditional DETR architecture for faster training convergence.

\section{Methodology}
In this section, we first introduce preliminaries and then discuss details of our method.
\subsection{Contrastive learning}
Contrastive learning has become popular due to
leading state-of-the-art results in different unsupervised representation learning tasks. The goal is to learn representations by contrasting positive pairs against negative pairs~\cite{hadsell2006dimensionality} and is used in various computer vision tasks, including data augmentation~\cite{oord2018representation}, or diverse scene generation~\cite{tian2020contrastive}. The main idea of contrastive learning is to move the similar pairs near and dissimilar pairs far in the embedding subspace.

\begin{figure*}
\centering
\includegraphics[width=0.6\textwidth]{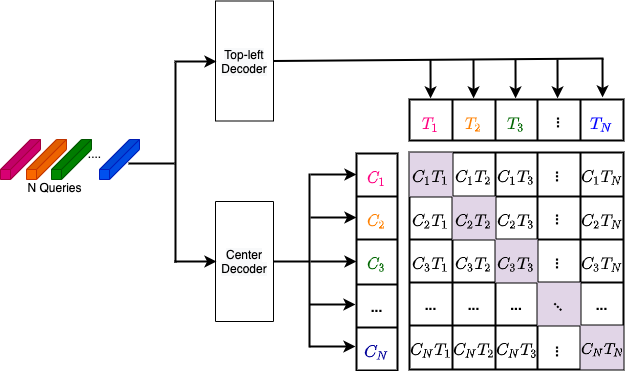}
\caption{Pair DETR utilizes contrastive learning to maximize the similarity between corresponding output embeddings (positive) of Center and Top-left decoders and minimizing the similarity to other output embeddings (negatives). }
\label{fig:contrastive}
\end{figure*}

In this work, we follow a similar approach to SimCLR framework~\cite{chen2020simple} for contrastive learning. SimCLR consists of four main components: a stochastic data augmentation method which generates positive pairs, an encoding network ($\mathrm{f}$) that extracts the embedding representation vectors from augmented samples, a small projector head ($\mathrm{g}$) that maps representations to the loss space, and a contrastive loss function that enforces more discriminative embedding representation between positive and negative pairs. Given a random mini-batch of N samples, SimCLR generates N positive pairs using the data
augmentation method. For all positive pairs, the remaining 2N-1 augmented samples are treated as negative examples. Let $\mathrm{h_i = f(x_i)}$ be the representations of all 2N
samples and $\mathrm{z_i = g(h_i)}$ be the projections of these representations. Let $\mathrm{sim(u, v) = u^Tv /||u|| ||v||}$ denote the dot product between $\ell_2$ normalized u and v. Then the loss function for a positive pair of examples (i, j) is defined as:
\begin{align}\label{eq1} 
	\ell_{i,j} = -\log \dfrac{\exp({\mathrm{sim(z_i,z_j)/\tau}})}{\Sigma^{2N}_{k=1}\mathbb{1}_{[k\neq {i}]}\exp({\mathrm{sim(z_i,z_j)/\tau}})}   ,
\end{align}
where $\mathbb{1} \in \{0, 1\}$ is an indicator function evaluating to
1 iff $\mathrm{\{k\neq {i}\}}$ and $\tau$ denotes a temperature parameter. The final loss which is also known as $\emph{NT-Xent}$~\cite {sohn2016improved,oord2018representation} considers the average of losses computed across all positive pairs, both (i, j) and (j, i), in a mini-batch~\cite{chen2020simple}.

\subsection{DETR}
The proposed approach follows detection transformer (DETR) and predicts all the objects at once without the need for anchor generation or NMS. DETR architecture consists of a CNN backbone, a transformer encoder, a transformer decoder, and object class and box position predictors. The goal of transformer encoder is to enhance the embeddings output from the CNN backbone. It is a stack of multiple encoder layers, where each layer mainly consists of a self-attention layer and a feed-forward layer. 

The transformer decoder is a stack of decoder layers. Each decoder layer, consists of three main components which are  self-attention, cross-attention, and feed-forward layers, respectively. Self-attention layer performs interactions between the embeddings, outputted from the previous decoder layer and is utilized to remove duplication in prediction. Cross-attention layer combines the embeddings output from the encoder to refine the decoder embeddings for enhancing class and box prediction. Conditional DETR utilizes conditional cross-attention mechanism and forms the query by concatenating the content query $\mathrm{c_q}$, outputting from decoder self-attention, and the spatial query $\mathrm{p_q}$.
\subsection{Pair DETR}

The proposed Pair DETR mechanism forms by creating a pair of points. There are evidences which show the superiority of using two corners (e.g. the top-left corner and
bottom-right corner) in terms of localization comparing to the other forms of object detection models which using center, width, height of bounding box representation. However, corner based models in general perform worse for the classification part [3]. One possible reason might be due to the fact that the corners contain less contextual information of the object inside the bounding box comparing to the center.
Pair DETR is to taking advantage of both of these models by considering a pair of the center of bounding box and one corner of the same bounding box. Having only one corner in addition to the center is enough since the bottom-right corner is reciprocal to the top-left one. In addition, to associate the corresponding points belonging to the same bounding box, the center and one corner probably have more similar contents comparing to two opposite corners.

\begin{table*}[]
\centering
\caption{Comparison between Pair DETR and previous DETR variants on COCO 2017 val. Our Pair DETR approach for all the backbones including high resolution (DC5-R50 and DC5-R101) and low resolution (R50 and R101) is at least 10× faster than the original DETR. In addition, Pair DETR performs almost 1.5$\times$ faster than conditional DETR. Pair DETR is empirically superior to other two single-scale DETR variants.}
\begin{tabular}{llllllll}
\hline
Model                            & \multicolumn{1}{l|}{\#epochs} & AP   & AP$_{50}$ & A$_{75}$ & AP$_{S}$ & AP$_{M}$ & AP$_L$ \\ \hline
DETR-R50                         & \multicolumn{1}{c|}{500}      & 42.0 & 62.4      & 44.2     & 20.5   & 45.8    & 61.1    \\
DETR-R50                         & \multicolumn{1}{c|}{50}       & 34.9 & 55.5      & 36.0     & 14.4   & 37.2    & 54.5    \\
Conditional DETR-R50             & \multicolumn{1}{c|}{50}       & 40.9 & 61.8      & 43.3     & 20.8   & 44.6    & 59.2    \\
Conditional DETR-R50             & \multicolumn{1}{c|}{75}       & 42.1 & 62.9      & 44.8     & 21.6   & 45.4    & 60.2    \\
Conditional DETR-R50             & \multicolumn{1}{c|}{108}      & 43.0 & 64.0      & 45.7     & 22.7   & 46.7    & 61.5    \\
Pair DETR-R50                    & \multicolumn{1}{c|}{50}       & 42.2 & 63.4      & 44.9     & 21.5   & 45.5    & 61.1    \\
Pair DETR-R50                    & \multicolumn{1}{c|}{75}       & 43.0 & 64.6      & 45.7     & 22.0   & 46.2    & 61.5    \\ \hline
DETR-DC5-R50                     & \multicolumn{1}{c|}{500}      & 43.3 & 63.1      & 45.9     & 22.5   & 47.3    & 61.1    \\
DETR-DC5-R50                     & \multicolumn{1}{c|}{50}       & 36.7 & 57.6      & 38.2     & 15.4   & 39.8    & 56.3    \\
Conditional DETR-DC5-R50         & \multicolumn{1}{c|}{50}       & 43.8 & 64.4      & 46.7     & 24.0   & 47.6    & 60.7    \\
Conditional DETR-DC5-R50         & \multicolumn{1}{c|}{75}       & 44.5 & 65.2      & 47.3     & 24.4   & 48.1    & 62.1    \\
Conditional DETR-DC5-R50         & \multicolumn{1}{c|}{108}      & 45.1 & 65.4      & 48.5     & 25.3   & 49.0    & 62.2    \\
Pair DETR-DC5-R50                & \multicolumn{1}{c|}{50}       & 44.6 & 65.4      & 47.5     & 24.5   & 48.1    & 61.9    \\
Pair DETR-DC5-R50                & \multicolumn{1}{c|}{75}       & 45.5 & 66.0      & 47.9     & 24.8   & 48.5    & 62.4    \\ \hline
DETR-R101                        & \multicolumn{1}{c|}{500}      & 43.5 & 63.8      & 46.4     & 21.9   & 48.0    & 61.8    \\
DETR-R101                        & \multicolumn{1}{c|}{50}       & 36.9 & 57.8      & 38.6     & 15.5   & 40.6    & 55.6    \\
Conditional DETR-R101            & \multicolumn{1}{c|}{50}       & 42.8 & 63.7      & 46.0     & 21.7   & 46.6    & 60.9    \\
Conditional DETR-R101            & \multicolumn{1}{c|}{75}       & 43.7 & 64.9      & 46.8     & 23.3   & 48.0    & 61.7    \\
Conditional DETR-R101            & \multicolumn{1}{c|}{108}      & 44.5 & 65.6      & 47.5     & 23.6   & 48.4    & 63.6    \\
Pair DETR-R101                   & \multicolumn{1}{c|}{50}       & 43.6 & 64.8      & 46.7     & 22.4   & 47.2    & 62.0    \\
Pair DETR-R101                   & \multicolumn{1}{c|}{75}       & 44.4 & 65.8      & 47.4     & 24.3   & 48.4    & 62.7    \\ \hline
DETR-DC5-R101                    & \multicolumn{1}{c|}{500}      & 44.9 & 64.7      & 47.7     & 23.7   & 49.5    & 62.3    \\
DETR-DC5-R101                    & \multicolumn{1}{c|}{50}       & 38.6 & 59.7      & 40.7     & 17.2   & 42.2    & 57.4    \\
Conditional DETR-DC5-R101        & \multicolumn{1}{c|}{50}       & 45.0 & 65.5      & 48.4     & 26.1   & 48.9    & 62.8    \\
Conditional DETR-DC5-R101        & \multicolumn{1}{c|}{75}       & 45.6 & 66.5      & 48.8     & 25.5   & 49.7    & 63.3    \\
Conditional DETR-DC5-R101        & \multicolumn{1}{c|}{108}      & 45.9 & 66.8      & 49.5     & 27.2   & 50.3    & 63.3    \\
Pair DETR-DC5-R101               & \multicolumn{1}{c|}{50}       & 45.7 & 66.4      & 48.9     & 26.4   & 49.3    & 63.6    \\
Pair DETR-DC5-R101               & \multicolumn{1}{c|}{75}       & 46.1 & 67.2      & 49.3     & 26.7   & 49.8    & 63.9    \\ \hline
Other Single-scale DETR variants &                               &      &           &          &        &         &         \\ \hline
Deformable DETR-R50-SS           & \multicolumn{1}{c|}{50}       & 39.4 & 59.6      & 42.3     & 20.6   & 43.0    & 55.5    \\
UP-DETR-R50                      & \multicolumn{1}{c|}{150}      & 40.5 & 60.8      & 42.6     & 19.0   & 44.4    & 60.0    \\
UP-DETR-R50                      & \multicolumn{1}{c|}{300}      & 42.8 & 63.0      & 45.3     & 20.8   & 47.1    & 61.7    \\ \hline
Deformable DETR-DC5-R50-SS       & \multicolumn{1}{c|}{50}       & 41.5 & 61.8      & 44.9     & 24.1   & 45.3    & 56.0    \\ \hline
\end{tabular}\label{comparetoDETR}
\end{table*}

Two parallel decoders are employed in Pair DETR. Feature maps generated from the encoder are used as common keys by the two decoders. Two decoders are designed to detect top-left and center points, respectively. Each decoder has its own classification and regression loss and both decoders are trained jointly. The same set of learnt positional encodings that are referred to as object queries are fed to the pair of decoders (i.e, top-left decoder and center decoder). Specifically, the first object query from this set which is fed to the top-left decoder will be specialized in detecting the top-left of first bounding box. In addition, this object query is fed to the center decoder to detect the center of corresponding bounding box.  These two embeddings from the two decoders are constructing the positive pair. Contrastive learning is utilized to keep consistent representation from outputs of two decoders. Given 2N queries and a positive pair, similar to~\cite{chen2020simple,radford2021learning}, we treat one pair as positive and the other 2N-1 queries as negative examples. The procedure is depicted in Figures~\ref{fig:method_overall} and ~\ref{fig:contrastive}.

\section{Results}
\subsection{Dataset}
Same as other work in the DETR family, we conduct experiments on COCO 2017 dataset~\cite{lin2014microsoft}. Our models are trained on the train set, and evaluated on the val set.

\begin{table*}[]
\centering
\caption{A comparison of Pair DETR to multi-scale and higher-resolution DETR variants. Pair DETR is a single-scale approach. However our approach when utilizing high resolution backbone such as DC5-R50 and DC5-R101 outperforms the two multi-scale and higher-resolution DETR variants. This is with using only single scale and without using high resolution encoder.}
\begin{tabular}{lc|llllll}
\hline
Model                     & \#epochs & AP   & AP$_{50}$ & AP$_{75}$ & AP$_S$ & AP$_M$ & AP$_L$ \\ \hline
Faster RCNN-FPN-R50~\cite{ren2015faster} & 36       & 40.2 & 61.0    & 43.8    & 24.2   & 43.5   & 52.0   \\
Faster RCNN-FPN-R50 & 108      & 42.0 & 62.1    & 45.5    & 26.6   & 45.5   & 53.4   \\
Deformable DETR-R50       & 50       & 43.8 & 62.6    & 47.7    & 26.4   & 47.1   & 58.0   \\
TSP-FCOS-R50~\cite{sun2021rethinking}              & 36       & 43.1 & 62.3    & 47.0    & 26.6   & 46.8   & 55.9   \\
TSP-RCNN-R50              & 36       & 43.8 & 63.3    & 48.3    & 28.6   & 46.9   & 55.7   \\
TSP-RCNN-R50              & 96       & 45.0 & 64.5    & 49.6    & 29.7   & 47.7   & 58.0   \\ \hline
Pair DETR-DC5-R50         & 50       & 44.6 & 65.4    & 47.5    & 24.5   & 48.1   & 61.9   \\
Pair DETR-DC5-R50         & 108      & 45.8 & 66.2    & 48.8    & 25.7   & 49.0   & 63.0   \\ \hline
Faster RCNN-FPN-R101      & 36       & 42.0 & 62.5    & 45.9    & 25.2   & 45.6   & 54.6   \\
Faster RCNN-FPN-R101      & 108      & 44.0 & 63.9    & 47.8    & 27.2   & 48.1   & 56.0   \\
TSP-FCOS-R101             & 36       & 44.4 & 63.8    & 48.2    & 27.7   & 48.6   & 57.3   \\
TSP-RCNN-R101             & 36       & 44.8 & 63.8    & 49.2    & 29.0   & 47.9   & 57.1   \\
TSP-RCNN-R101             & 96       & 46.5 & 66.0    & 51.2    & 29.9   & 49.7   & 59.2   \\ \hline
Pair DETR-DC5-R101        & 50       & 45.7 & 66.4    & 48.9    & 26.4   & 49.3   & 63.6   \\
Pair DETR-DC5-R101        & 108      & 46.6 & 67.5    & 49.8    & 27.4   & 50.6   & 64.4   \\ \hline
\end{tabular}\label{othercomparison}
\end{table*}

\subsection{Implementation details}
Pair DETR is based on conditional DETR architecture~\cite{meng2021conditional} and consists of the CNN backbone, transformer encoder, two transformer decoders,  prediction feed-forward networks (FFNs) following each decoder layer (the last decoder layer and the 5 internal decoder layers) with parameters shared among the 6 prediction FFNs. Classification network (FFN) is shared between the two decoders (center and top-left decoders) while the regression network is separated. The hyperparameters are the same as DETR and conditional DETR. Conditional spatial embeddings are employed as the spatial queries for conditional multi-head cross-attention and the combination of the spatial query (key) and the content query (key) is done through concatenation other than addition. 
In the first cross-attention layer there are no decoder content embeddings, we make simple changes based on the DETR implementation: concatenate the positional embedding predicted from the object query (the positional embedding) into the original query (key).

The prediction unit for spatial queries is an FFN and consists of learnable linear projection + ReLU + learnable linear projection: $ \mathrm {s =FFN(o_q)}$. The 2D coordinates are normalized by the sigmoid function to use for forming conditional spatial query.
 
We utilized the Hungarian algorithm to find an optimal bipartite matching between predicted and the ground-truth object following DETR~\cite{carion2020end} and then form the loss function for computing and back-propagate the gradients. We follow Deformable DETR~\cite{zhu2020deformable} to formulate the loss: the same loss function with 300 object queries, the same matching cost function, and the same trade-off parameters; The classification loss function is focal loss~\cite{lin2017focal}, and the box regression loss (including L1 and GIoU~\cite{rezatofighi2019generalized} loss) is the same as DETR~\cite{carion2020end}. Each decoder has its own classification and regression loss. Contrastive loss is also utilized between two decoders.

\begin{figure*}
\begin{tabular}{cccc}
\includegraphics[width=0.23\textwidth]{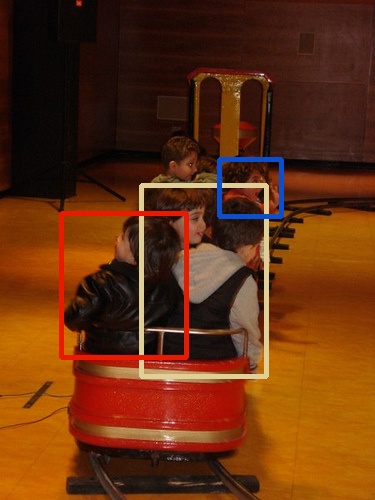} & \includegraphics[width=0.23\textwidth]{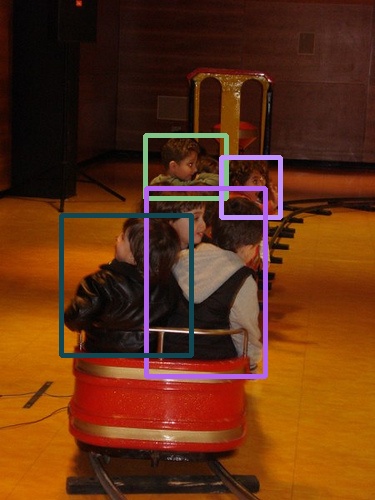} & \includegraphics[width=0.23\textwidth]{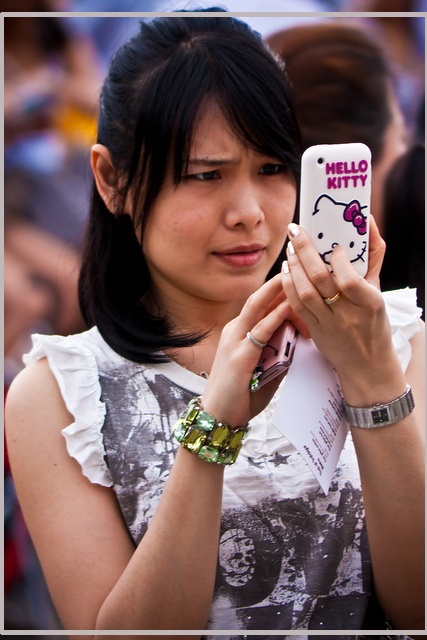} & \includegraphics[width=0.23\textwidth]{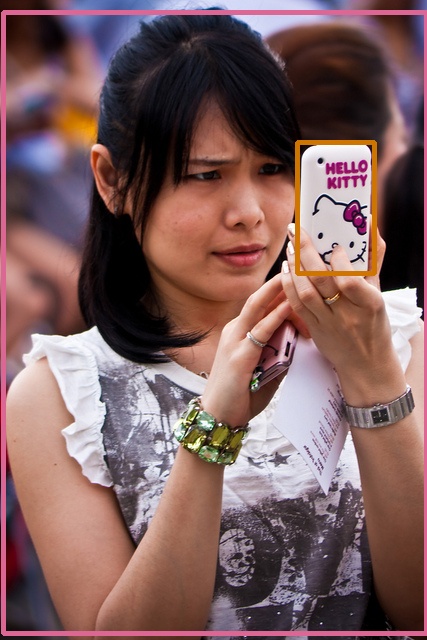} \\
(a) Conditional DETR & (b) Ours & (c) Conditional DETR & (d) Ours \\
\includegraphics[width=0.23\textwidth]{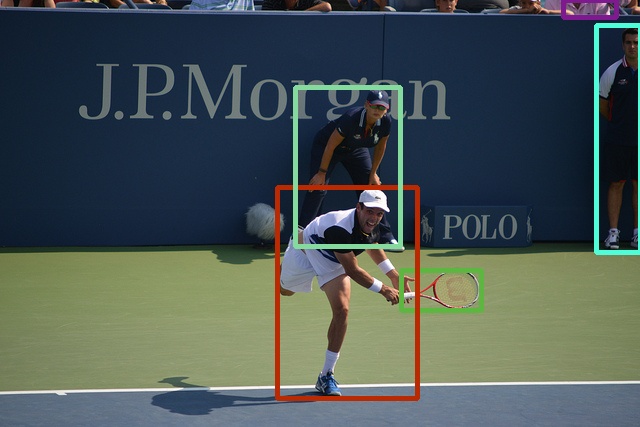} & \includegraphics[width=0.23\textwidth]{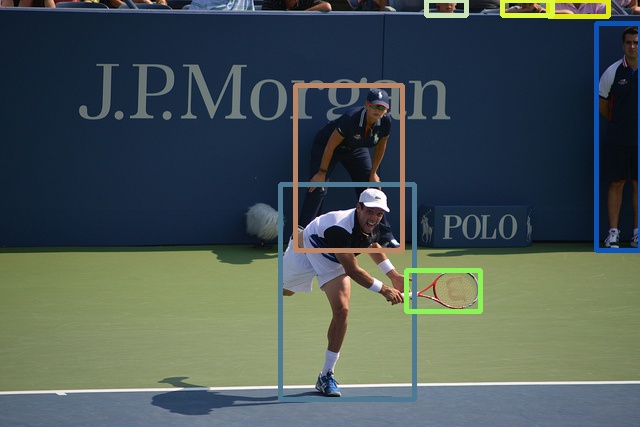} & \includegraphics[width=0.23\textwidth]{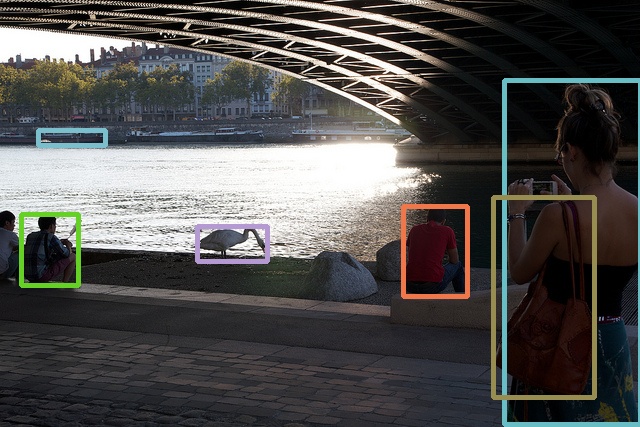} & \includegraphics[width=0.23\textwidth]{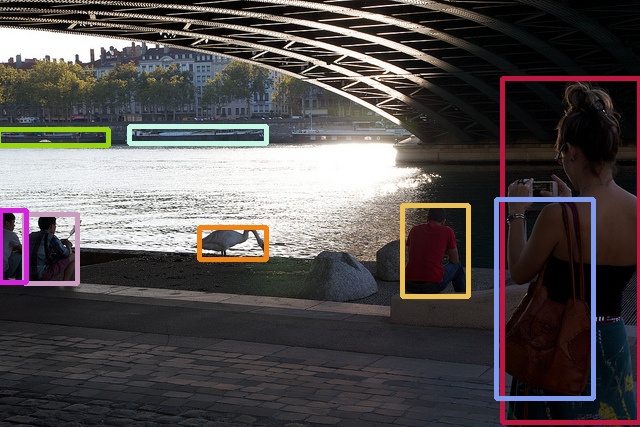} \\
(e) Conditional DETR & (f) Ours  & (g) Conditional DETR & (h) Ours \\
\end{tabular}
\caption{{Visual comparison on COCO validation set.} Comparing to conditional DETR~\cite{meng2021conditional}, Pair DETR shows superior performance in detecting objects blending in the background (\ref{visualizatoncoco}a vs. b, \ref{visualizatoncoco}g vs. h), locating smaller objects contained within larger objects (\ref{visualizatoncoco}c vs. d), as well as detecting objects around the boundaries of images (\ref{visualizatoncoco}e vs. f, \ref{visualizatoncoco}g vs. h).}
\label{visualizatoncoco}
\end{figure*}

Training is performed following the DETR training protocol~\cite{carion2020end}. The backbone is the pretrained model on ImageNet from torchvision with batchnorm layers fixed, and the transformer parameters are initialized using the Xavier initialization scheme~\cite{glorot2010understanding}. The weight decay is set to be $10^{-4}$. The AdamW~\cite{loshchilov2018fixing} optimizer is employed.  The learning rates for the backbone and the transformer are initially set to be $10^{-5}$ and $10^{-4}$, respectively. The dropout rate in transformer is 0.1. The learning rate is dropped by a factor of 10 after 40 epochs for 50 training epochs, after 60 epochs for 75 training epochs, and after 80 epochs for 108 training epochs. We use data augmentation following DETR scheme: resizing the input images whose shorter side is between 480 to 800 pixels while the longer side is by at most 1333 pixels; randomly crop the image such that a training image is cropped with probability 0.5 to a random rectangular patch.
We use the standard COCO to evaluate and report the average precision (AP), and the AP scores at 0.50, 0.75 and for the small, medium, and large objects.

\subsection{Comparison to DETR}

We compare the proposed Pair DETR to the original DETR~\cite{carion2020end} and conditional DETR~\cite{meng2021conditional}. We follow~\cite{carion2020end} and report the results over four backbones: ResNet-50~\cite{he2016deep}, ResNet-101, and their 
high resolution extensions DC5-ResNet-50 and DC5-ResNet-101.

The corresponding DETR models are named as DETR-R50, DETR-R101, DETR-DC5-R50, and DETR-DC5-
R101, respectively. Our models are named as Pair DETR-R50, Pair DETR-R101, Pair DETR-DC5-R50, and Pair DETR-DC5-R101, respectively. 

Table \ref{comparetoDETR} presents the results from DETR, conditional DETR, and Pair DETR. DETR with 50 training epochs performs much worse than 500 training epochs. Conditional DETR with 50 training epochs for R50 and R101 as the backbones performs slightly worse than DETR with 500 training epochs. However Pair DETR (backbone R50 and R101) performs slightly better than DETR with 500 training epochs. In addition, Pair DETR with 50 training epochs for DC5-R50 and DC5-R101 could outperform DETR with 500 training epochs by a bigger margin. Pair DETR for the three backbones with 50 training epochs performs slightly better than conditional DETR with 75 training epochs (the only exception is backbone R101). Figure \ref{visualizatoncoco} visualizes the detection performance of Pair vs. conditional DETR on COCO dataset with 50 training epochs. In addition, Pair DETR with 75 training epochs outperform conditional DETR with 75 training epochs with a larger margin and for all the four backbones. In summary, training Pair DETR for all the backbones is at least 10$\times$ faster than the original DETR. In addition, training Pair DETR for all the backbones is 1.5$\times$ faster than corresponding conditional DETR.

\newpage
\begin{table*}[]
\centering
\caption{A comparison of different settings for Pair DETR. We investigate the effect of adding each component to the proposed Pair DETR. Addition of contrastive learning showed effective improvement. In addition, localization of bounding boxes based on two points is more accurate compared to regressing it directly from the center.}
\begin{tabular}{lccc|llllll}
\hline
Model         & Cont. learning & w, h         & Pairs coordinates  & AP   & AP$_{50}$ & AP$_{75}$ & AP$_S$ & AP$_M$ & AP$_L$ \\ \hline
Pair DETR-R50 (a) &                & $\checkmark$ &                        & 41.1 & 62.1    & 43.5    & 20.9   & 44.8   & 59.3   \\
Pair DETR-R50 (b) & $\checkmark$   & $\checkmark$ &                         & 41.8 & 63.3    & 44.1    & 21.1   & 45.2   & 60.7   \\
Pair DETR-R50 (c) & $\checkmark$   &              & $\checkmark$            & 42.1 & 63.6    & 44.5    & 21.3   & 45.4   & 61.0   \\
Pair DETR-R50 (d) & $\checkmark$   & $\checkmark$ & $\checkmark$            & 42.2 & 63.8    & 44.5    & 21.5   & 45.5   & 61.1   \\ \hline
\end{tabular}
\label{Tableablation1}
\end{table*}

\begin{table*}[]
\centering
\caption{A comparison of different matching strategies for Pair DETR.}
\begin{tabular}{lcccc|llllll}
\hline
Model    & head sep.    & head sh.  & Tl cl. cost  & Cent. cl. cost & AP   & AP$_{50}$ & A$_{75}$ & AP$_S$ & AP$_M$ & AP$_L$ \\ \hline
Pair DETR-R50 (1) & $\checkmark$ &              & $\checkmark$ & $\checkmark$   & 41.4 & 62.5      & 43.8     & 21.0   & 44.9   & 59.8   \\
Pair DETR-R50 (2) &              & $\checkmark$ & $\checkmark$ & $\checkmark$   & 42.0 & 63.4      & 44.2     & 21.3   & 45.3   & 60.7   \\
Pair DETR-R50 (3) &              & $\checkmark$ &              & $\checkmark$   & 42.2 & 63.8      & 44.5     & 21.5   & 45.5   & 61.1  
\\ \hline
\end{tabular}
\label{ablation2table}
\end{table*}
In addition, we report the results of single-scale DETR
extensions: UP-DETR~\cite{dai2021up} and deformable DETR-SS~\cite{zhu2020deformable} in Table \ref{comparetoDETR}. Our results over R50 and DC5-R50 are better than deformable DETR-SS: 42.2 vs. 39.4 and 44.6 vs. 41.5. 
The comparison might not be fully fair as for example parameter and computation complexities are more different in these extensions, but it implies that the conditional cross-attention mechanism is beneficial. Compared to UP-DETR-R50, our results with fewer training epochs are obviously better.

Overall, some people may take the higher AP in Table~\ref{comparetoDETR} as granted as Pair DETR obviously has more parameters than some of the other methods, e.g., the original DETR and the Conditional DETR, to fit the data. However, in addition to accuracy, it is important to read it from another perspective: Pair DETR demonstrates significantly faster convergence than its counterparts, although having a larger size of model.

\subsection{Comparison to multi-scale and higher-resolution DETR variants}

It has been shown in prior work that conventional wisdom in computer vision such as embracing higher resolution and multi scale can also boost object detection accuracy for DETRs. 
As they are techniques orthogonal to this paper, it is possible for future extension of Pair DETR also benefits from multi-scale attention and/or higher-resolution encoder. Although we did not perform these implementations, Table \ref{othercomparison} lists the performance of these methods and Pair DETR using higher resolution backbones as a comparison. It can be seen that our methods are on par or better than these recent work. 



\subsection{Ablation}
\textbf{Effect of contrastive loss and regressing bounding box.} We consider four different models to study the effect of different components of the Pair DETR. All models are trained for 50 epochs and the classification head is shared between two decoders. However, each decoder has it own regression head. In the setting (a) contrastive learning is not employed and also (w, h) is used to regress the bounding box from its center. This model performance is very similar to the conditional DETR. 

In second setting (b), the model is jointly trained using contrastive loss and the classification and regression loss functions. Adding contrastive loss function causes the decoders embedding subspace to be more discriminative and consequently the detection performance is improved. In the third setting (c), we utilize the coordinates of two points to measure width and height instead of regressing them directly from the center points. Therefore the conversion function becomes:
\begin{align}\label{coordinates}
	  \mathrm{T(w,h)} = \Big(\mathrm{(x_{tl}-x_{c})} \times 2 , \mathrm{(y_{tl}-y_{c})} \times 2 \Big),
\end{align}
where $(\mathrm{x_{tl}, y_{tl}})$ and $(\mathrm{x_{c}, y_{c}})$ are the coordinates of top-left and center points, respectively. This conversion makes the model to consider two important points instead of relying on one point (center) and it has a positive impact on the detection performance. In the last setting (d), width and height are extracted using Eq.\ref{coordinates} and also predicted independently from center decoder. The conversion function for width and height is taking average of them as follows:
\begin{align}
	  \mathrm{T(w,h)} = \Big(\dfrac{\mathrm{(x_{tl}-x_{c})} \times 2 + \mathrm{w_c}}{2} , \dfrac{\mathrm{(y_{tl}-y_{c})} \times 2 + \mathrm{h_c}}{2} \Big),
\end{align}\label{coordinates_2}
where $(\mathrm{x_{tl}, y_{tl}})$ and $(\mathrm{x_{c}, y_{c}})$ are the coordinates of top-left and center points, respectively. $(\mathrm{w_{c}, h_{c}})$ is the predicted width and height from the center decoder. This setting outperforms the other settings as it is presented in Table \ref{Tableablation1}. 
Here we take the average of the calculated (w,h) from the pair coordinates as well as the predicted width and height from the center decoder. 

\textbf{Strategies for Hungarian matching.}
We evaluate different strategies for Hungarian matching. As it is explained in original DETR paper, after the bounding box prediction step, they are selected in a way to minimize the classification and regression costs. In this experiment, we are interested to explore the effect of classification cost on Hungarian matching and model performance. Three different settings are considered and in all of them the model is trained with 50 training epochs. In the first setting (1), each decoder has its own classification head. Moreover, Top-left decoder has the same role as center decoder in classification cost of Hungarian matching. This model performs worst among all the three settings. In the second setting (2), the classification head is shared between two decoders, but still the classification cost of Top-left decoder plays equal role and is considered in matching. In the third setting (3), the classification head is shared and only the classification cost of center decoder is considered for matching. As it is shown in Table \ref{ablation2table}, this model outperforms the other two settings. One possible reason for this can be due to the fact that the center points contain more representative features related to the content (class id) of the bounding box compared to extreme points or corner points. Therefore this setting is selected for our proposed Pair DETR model. 

\section{Conclusions}
In this work, we introduce a simple approach for considering extra point in DETR which detects objects using a pair, including one center keypoint and one corner (Top-left). For this purpose, one additional decoder is employed.
Using this approach, we leverage more than one keypoint for the localization of a bounding box. In addition, we utilize a contrastive learning approach as an auxiliary task to make the output embeddings of two decoders to be more discriminative. Although the presented implementation is based upon conditional DETR, the simplicity of the idea allows it easily extended to other DETR family detectors e.g., Deformable DETR~\cite{zhu2020deformable} in the future, expectably boosting the performance of the base detector by a considerable margin.

{
    \small
    \bibliographystyle{ieee_fullname}
    \bibliography{main}

\begin{thebibliography}{10}\itemsep=-1pt

\bibitem{carion2020end}
Nicolas Carion, Francisco Massa, Gabriel Synnaeve, Nicolas Usunier, Alexander
  Kirillov, and Sergey Zagoruyko.
\newblock End-to-end object detection with transformers.
\newblock In {\em European Conference on Computer Vision}, pages 213--229.
  Springer, 2020.

\bibitem{chen2020simple}
Ting Chen, Simon Kornblith, Mohammad Norouzi, and Geoffrey Hinton.
\newblock A simple framework for contrastive learning of visual
  representations.
\newblock In {\em International conference on machine learning}, pages
  1597--1607. PMLR, 2020.

\bibitem{chen2020reppoints}
Yihong Chen, Zheng Zhang, Yue Cao, Liwei Wang, Stephen Lin, and Han Hu.
\newblock Reppoints v2: Verification meets regression for object detection.
\newblock {\em Advances in Neural Information Processing Systems}, 33, 2020.

\bibitem{dai2016r}
Jifeng Dai, Yi Li, Kaiming He, and Jian Sun.
\newblock R-fcn: Object detection via region-based fully convolutional
  networks.
\newblock In {\em Advances in neural information processing systems}, pages
  379--387, 2016.

\bibitem{dai2021up}
Zhigang Dai, Bolun Cai, Yugeng Lin, and Junying Chen.
\newblock Up-detr: Unsupervised pre-training for object detection with
  transformers.
\newblock In {\em Proceedings of the IEEE/CVF Conference on Computer Vision and
  Pattern Recognition}, pages 1601--1610, 2021.

\bibitem{glorot2010understanding}
Xavier Glorot and Yoshua Bengio.
\newblock Understanding the difficulty of training deep feedforward neural
  networks.
\newblock In {\em Proceedings of the thirteenth international conference on
  artificial intelligence and statistics}, pages 249--256. JMLR Workshop and
  Conference Proceedings, 2010.

\bibitem{hadsell2006dimensionality}
Raia Hadsell, Sumit Chopra, and Yann LeCun.
\newblock Dimensionality reduction by learning an invariant mapping.
\newblock In {\em 2006 IEEE Computer Society Conference on Computer Vision and
  Pattern Recognition (CVPR'06)}, volume~2, pages 1735--1742. IEEE, 2006.

\bibitem{he2016deep}
Kaiming He, Xiangyu Zhang, Shaoqing Ren, and Jian Sun.
\newblock Deep residual learning for image recognition.
\newblock In {\em Proceedings of the IEEE conference on computer vision and
  pattern recognition}, pages 770--778, 2016.

\bibitem{law2018cornernet}
Hei Law and Jia Deng.
\newblock Cornernet: Detecting objects as paired keypoints.
\newblock In {\em Proceedings of the European conference on computer vision
  (ECCV)}, pages 734--750, 2018.

\bibitem{lin2017focal}
Tsung-Yi Lin, Priya Goyal, Ross Girshick, Kaiming He, and Piotr Doll{\'a}r.
\newblock Focal loss for dense object detection.
\newblock In {\em Proceedings of the IEEE international conference on computer
  vision}, pages 2980--2988, 2017.

\bibitem{lin2014microsoft}
Tsung-Yi Lin, Michael Maire, Serge Belongie, James Hays, Pietro Perona, Deva
  Ramanan, Piotr Doll{\'a}r, and C~Lawrence Zitnick.
\newblock Microsoft coco: Common objects in context.
\newblock In {\em European conference on computer vision}, pages 740--755.
  Springer, 2014.

\bibitem{liu2016ssd}
Wei Liu, Dragomir Anguelov, Dumitru Erhan, Christian Szegedy, Scott Reed,
  Cheng-Yang Fu, and Alexander~C Berg.
\newblock Ssd: Single shot multibox detector.
\newblock In {\em European conference on computer vision}, pages 21--37.
  Springer, 2016.

\bibitem{loshchilov2018fixing}
Ilya Loshchilov and Frank Hutter.
\newblock Fixing weight decay regularization in adam.
\newblock 2018.

\bibitem{meng2021conditional}
Depu Meng, Xiaokang Chen, Zejia Fan, Gang Zeng, Houqiang Li, Yuhui Yuan, Lei
  Sun, and Jingdong Wang.
\newblock Conditional detr for fast training convergence.
\newblock In {\em Proceedings of the IEEE/CVF International Conference on
  Computer Vision}, pages 3651--3660, 2021.

\bibitem{oord2018representation}
Aaron van~den Oord, Yazhe Li, and Oriol Vinyals.
\newblock Representation learning with contrastive predictive coding.
\newblock {\em arXiv preprint arXiv:1807.03748}, 2018.

\bibitem{radford2021learning}
Alec Radford, Jong~Wook Kim, Chris Hallacy, Aditya Ramesh, Gabriel Goh,
  Sandhini Agarwal, Girish Sastry, Amanda Askell, Pamela Mishkin, Jack Clark,
  et~al.
\newblock Learning transferable visual models from natural language
  supervision.
\newblock {\em arXiv preprint arXiv:2103.00020}, 2021.

\bibitem{redmon2016you}
Joseph Redmon, Santosh Divvala, Ross Girshick, and Ali Farhadi.
\newblock You only look once: Unified, real-time object detection.
\newblock In {\em Proceedings of the IEEE conference on computer vision and
  pattern recognition}, pages 779--788, 2016.

\bibitem{redmon2017yolo9000}
Joseph Redmon and Ali Farhadi.
\newblock Yolo9000: better, faster, stronger.
\newblock In {\em Proceedings of the IEEE conference on computer vision and
  pattern recognition}, pages 7263--7271, 2017.

\bibitem{ren2015faster}
Shaoqing Ren, Kaiming He, Ross Girshick, and Jian Sun.
\newblock Faster r-cnn: Towards real-time object detection with region proposal
  networks.
\newblock {\em Advances in neural information processing systems}, 28:91--99,
  2015.

\bibitem{rezatofighi2019generalized}
Hamid Rezatofighi, Nathan Tsoi, JunYoung Gwak, Amir Sadeghian, Ian Reid, and
  Silvio Savarese.
\newblock Generalized intersection over union: A metric and a loss for bounding
  box regression.
\newblock In {\em Proceedings of the IEEE/CVF Conference on Computer Vision and
  Pattern Recognition}, pages 658--666, 2019.

\bibitem{sohn2016improved}
Kihyuk Sohn.
\newblock Improved deep metric learning with multi-class n-pair loss objective.
\newblock In {\em Advances in neural information processing systems}, pages
  1857--1865, 2016.

\bibitem{sun2021rethinking}
Zhiqing Sun, Shengcao Cao, Yiming Yang, and Kris~M Kitani.
\newblock Rethinking transformer-based set prediction for object detection.
\newblock In {\em Proceedings of the IEEE/CVF International Conference on
  Computer Vision}, pages 3611--3620, 2021.

\bibitem{tian2020contrastive}
Yonglong Tian, Dilip Krishnan, and Phillip Isola.
\newblock Contrastive multiview coding.
\newblock In {\em Computer Vision--ECCV 2020: 16th European Conference,
  Glasgow, UK, August 23--28, 2020, Proceedings, Part XI 16}, pages 776--794.
  Springer, 2020.

\bibitem{tian2019fcos}
Zhi Tian, Chunhua Shen, Hao Chen, and Tong He.
\newblock Fcos: Fully convolutional one-stage object detection.
\newblock In {\em Proceedings of the IEEE/CVF international conference on
  computer vision}, pages 9627--9636, 2019.

\bibitem{tychsen2017denet}
Lachlan Tychsen-Smith and Lars Petersson.
\newblock Denet: Scalable real-time object detection with directed sparse
  sampling.
\newblock In {\em Proceedings of the IEEE international conference on computer
  vision}, pages 428--436, 2017.

\bibitem{vaswani2017attention}
Ashish Vaswani, Noam Shazeer, Niki Parmar, Jakob Uszkoreit, Llion Jones,
  Aidan~N Gomez, {\L}ukasz Kaiser, and Illia Polosukhin.
\newblock Attention is all you need.
\newblock In {\em Advances in neural information processing systems}, pages
  5998--6008, 2017.

\bibitem{zheng2020end}
Minghang Zheng, Peng Gao, Xiaogang Wang, Hongsheng Li, and Hao Dong.
\newblock End-to-end object detection with adaptive clustering transformer.
\newblock {\em arXiv preprint arXiv:2011.09315}, 2020.

\bibitem{zhou2019objects}
Xingyi Zhou, Dequan Wang, and Philipp Kr{\"a}henb{\"u}hl.
\newblock Objects as points.
\newblock {\em arXiv preprint arXiv:1904.07850}, 2019.

\bibitem{zhou2019bottom}
Xingyi Zhou, Jiacheng Zhuo, and Philipp Krahenbuhl.
\newblock Bottom-up object detection by grouping extreme and center points.
\newblock In {\em Proceedings of the IEEE/CVF Conference on Computer Vision and
  Pattern Recognition}, pages 850--859, 2019.

\bibitem{zhu2020deformable}
Xizhou Zhu, Weijie Su, Lewei Lu, Bin Li, Xiaogang Wang, and Jifeng Dai.
\newblock Deformable detr: Deformable transformers for end-to-end object
  detection.
\newblock {\em arXiv preprint arXiv:2010.04159}, 2020.

\end{thebibliography}
}



\end{document}